\documentclass{IEEEtran}
\usepackage{cite}
\usepackage{amsmath,amssymb,amsfonts}
\usepackage{algorithm}
\usepackage{algorithmicx}
\usepackage{algpseudocode}
\usepackage{graphicx}
\usepackage{textcomp}
\usepackage{xcolor}
\usepackage{subfigure}
\usepackage{color}
\usepackage{multirow}
\usepackage{booktabs}
\usepackage{pifont}
\usepackage{amsthm}
\usepackage{array}
\usepackage{stfloats} 
\usepackage{CJKutf8}
\usepackage{xspace}

\usepackage[capitalize]{cleveref}

\def\BibTeX{{\rm B\kern-.05em{\sc i\kern-.025em b}\kern-.08em
    T\kern-.1667em\lower.7ex\hbox{E}\kern-.125emX}}
\begin{document}

\title{FedModule: A Modular Federated Learning Framework}

\author{Chuyi Chen, Zhe Zhang, Yanchao Zhao\\
	\IEEEauthorblockA{College of Computer Science and Technology, Nanjing University of Aeronautics and Astronautics, China\\
 Email: yczhao@nuaa.edu.cn
		}
}

\maketitle

\begin{abstract}
Federated learning (FL) has been widely adopted across various applications, such as healthcare, finance, and smart cities. However, as experimental scenarios become more complex, existing FL frameworks and benchmarks have struggled to keep pace. This paper introduces FedModule\footnote{code is available at https://github.com/NUAA-SmartSensing/async-FL}, a flexible and extensible FL experimental framework that has been open-sourced to support diverse FL paradigms and provide comprehensive benchmarks for complex experimental scenarios. FedModule adheres to the "one code, all scenarios" principle and employs a modular design that breaks the FL process into individual components, allowing for the seamless integration of different FL paradigms. The framework supports synchronous, asynchronous, and personalized federated learning, with over 20 implemented algorithms. Experiments conducted on public datasets demonstrate the flexibility and extensibility of FedModule. The framework offers multiple execution modes—including linear, threaded, process-based, and distributed—enabling users to tailor their setups to various experimental needs. Additionally, FedModule provides extensive logging and testing capabilities, which facilitate detailed performance analysis of FL algorithms. Comparative evaluations against existing FL toolkits, such as TensorFlow Federated, PySyft, Flower, and FLGo, highlight FedModule's superior scalability, flexibility, and comprehensive benchmark support. By addressing the limitations of current FL frameworks, FedModule marks a significant advancement in FL experimentation, providing researchers and practitioners with a robust tool for developing and evaluating FL algorithms across a wide range of scenarios.
\end{abstract}

\section{Introduction}

Nowadays, Federated Learning (FL)\cite{mcmahan2017communication,nishio2019client} has been widely used in various applications, such as healthcare, finance, and smart cities\cite{li2020federated,cheng2021secureboost,ramu2022federated}. In FL, the data are distributed among multiple clients, and the model is trained on the data of the clients without uploading the data to the server. The server aggregates the model updates from the clients and updates the global model. Massive research has been conducted to improve the performance of FL, such as communication-efficient algorithms\cite{xie2019asynchronous,chen2024fedvc}, secure aggregation\cite{bonawitz2016secure}, model personalization\cite{pfedme}, and client heterogeneity\cite{nishio2019client}. However, as the depth and width of FL research methods evolve, experimental scenarios become increasingly complex, yet the associated experimental frameworks and benchmarks have not kept pace. The lack of a unified experimental framework and benchmark makes it difficult to compare the performance of different FL algorithms and reproduce the results of existing algorithms\cite{wang2023flgo, beutel2020flower}. This has become a bottleneck in the development of FL research.

Recently, several FL frameworks have been proposed to address this issue\cite{bonawitz2019TFF, beutel2020flower, wang2023flgo, ryffel2018syft}. For example, TensorFlow Federated(TFF)\cite{bonawitz2019TFF} provides a simulation environment for FL algorithms, and PySyft and Flower provide a distributed computing environment for FL. However, these frameworks are designed for specific scenarios and lack flexibility. For example, TFF is designed for FL algorithms based on synchronous federated learning, and PySyft is designed for FL algorithms with differential privacy. These frameworks are not suitable for comparing the performance of FL algorithms in different scenarios. For instance, when comparing synchronous algorithms with asynchronous algorithms, or synchronous algorithms with personalized algorithms, the existing frameworks face significant challenges in implementing these algorithms. Furthermore, they lack the necessary benchmarks for conducting experiments. Therefore, it is necessary to design a flexible and extensible FL experimental framework that supports various federated learning paradigms and provides a rich set of benchmarks to address complex and varied experimental scenarios.

\begin{figure}[tbp]
    \centering
    \includegraphics[width=0.8\linewidth]{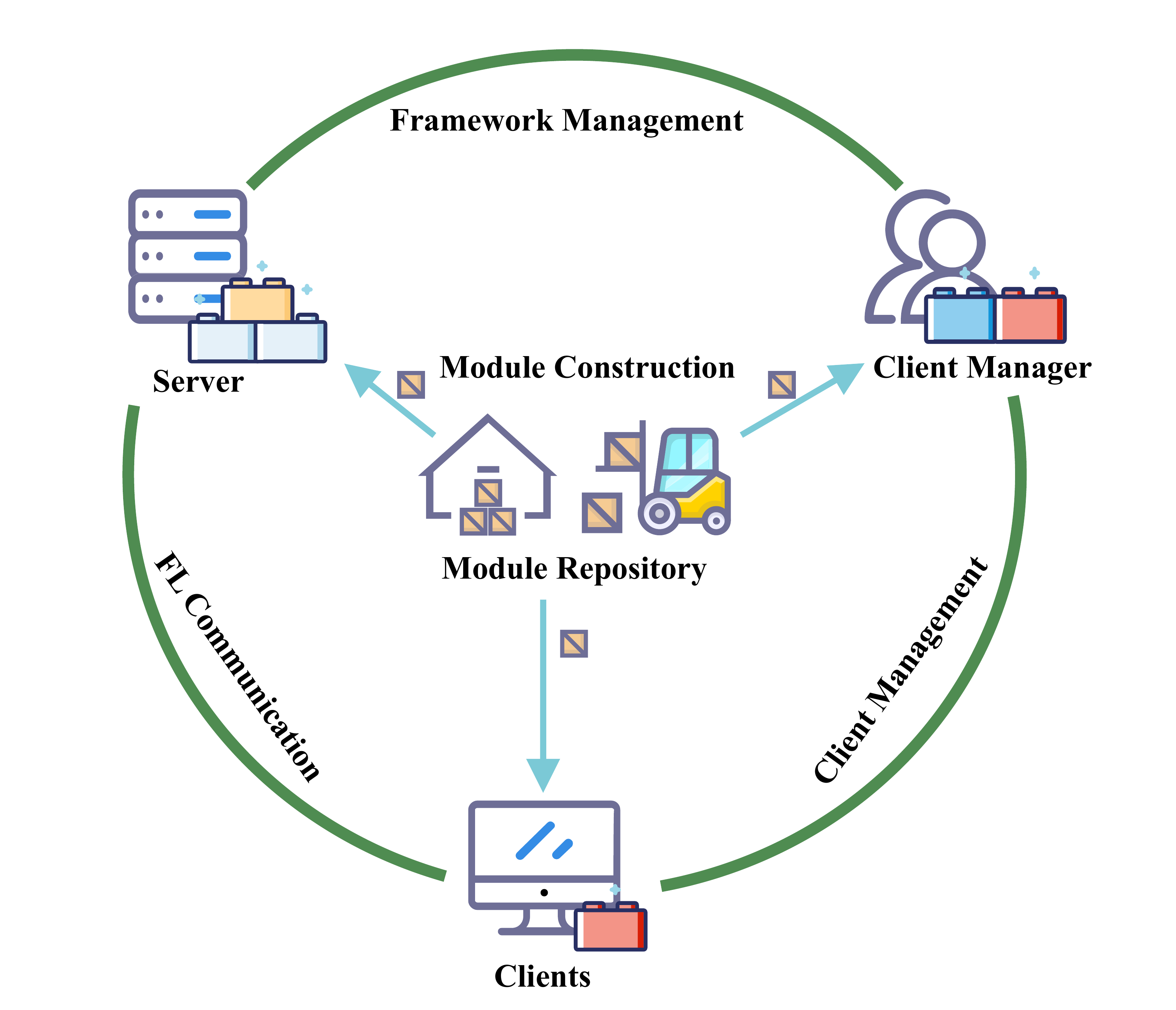}
    \caption{The conceptual diagram of FedModule illustrates how the framework conducts FL experiments. FedModule selects modules from the Module Repository via a process called Module Construction to build the three main roles in federated learning: clients, server, and client manager.}
    \label{fig:chap1}
\end{figure}

Thus, we propose our framework—FedModule. FedModule decomposes Federated Learning into individual modules, enabling it to seamlessly expand to support new paradigms and benchmarks. As illustrated in Fig. \labelcref{fig:chap1}, FedModule assembles various roles according to user requirements, with each module being replaceable and extendable to accommodate new federated learning paradigms. For instance, if a user wishes to implement round-robin scheduling instead of random scheduling, they can simply specify the round-robin scheduling algorithm module in the configuration, and the server will utilize the designated algorithm. Subsequently, when FedModule assembles the server, it integrates this module, enabling the server to apply the specified scheduling algorithm during the training process.

Furthermore, the FedModule framework adheres to the "one code, all scenarios" principle, enabling users to switch seamlessly between different execution modes and experimental scenarios by implementing their required code only once. Our main contributions are as follows:

\begin{itemize}
    \item We developed a modular federated learning framework with an extensive set of modules that can be combined to offer various execution modes and a wide range of benchmarks.
    \item FedModule encompasses multiple federated learning paradigms, including synchronous, asynchronous, and personalized federated learning, with over 20 implemented algorithms supporting these paradigms.
    \item We assessed the performance of FedModule across multiple benchmarks, demonstrating its flexibility and extensibility. Moreover, we tested the execution modes, and the results validated the effectiveness of each mode, offering users diverse configuration options for their experiments.
\end{itemize}

\section{Related Work}
\begin{figure*}[htbp]
    \centering
    \includegraphics[width=0.9\linewidth]{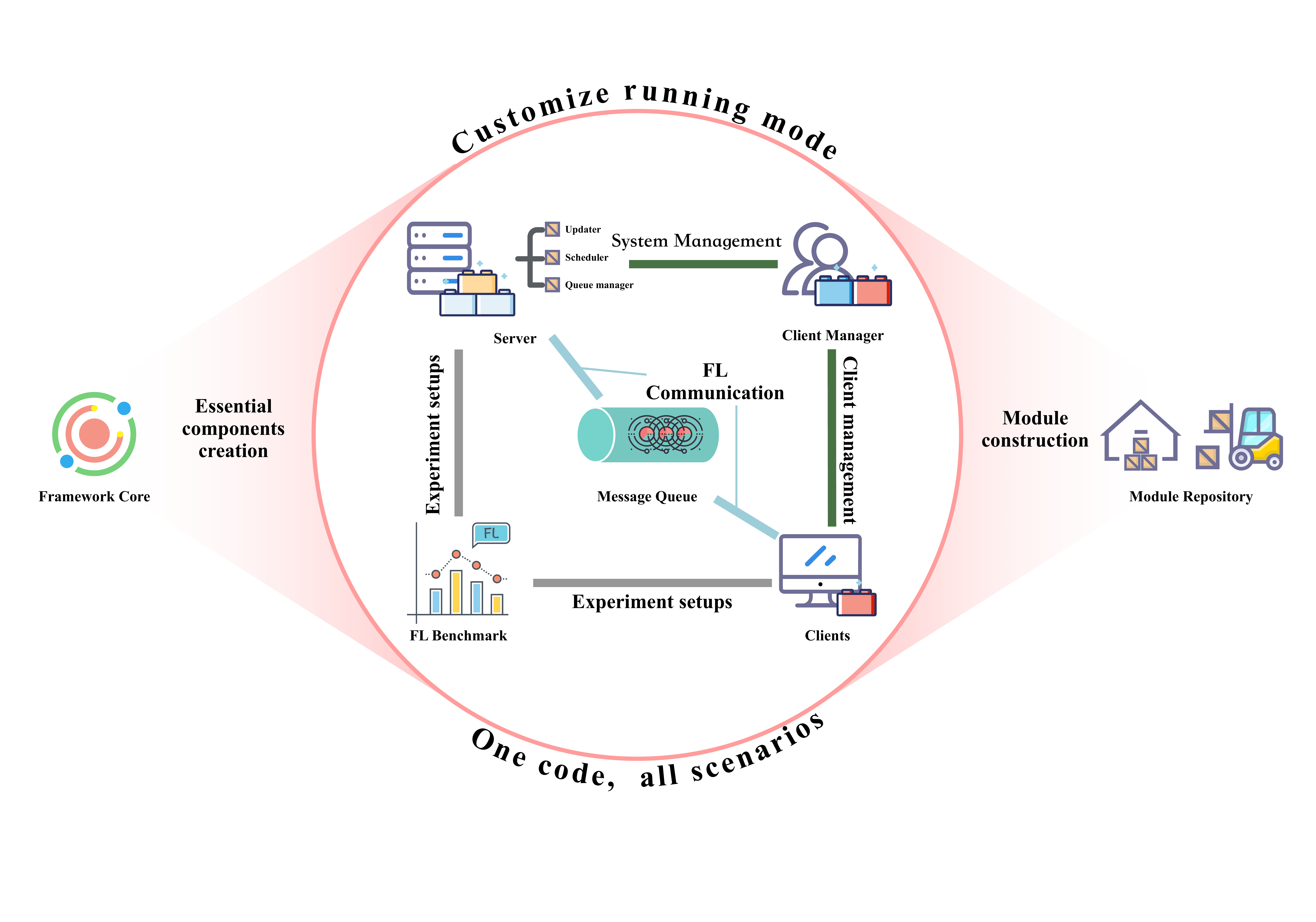}
    \caption{System Overview}
    \label{fig:system_overview}
\end{figure*}
Since Google introduced Federated Learning in 2017\cite{mcmahan2017communication}, numerous Federated Learning algorithms have been proposed. However, the development of frameworks suitable for related experiments has not progressed as quickly, with only a limited number of frameworks being introduced. TensorFlow Federated(TFF)\cite{bonawitz2019TFF} is a widely used FL framework that provides a simulation environment for FL algorithms. It is principally intended to simulate the training process of a limited number of homogeneous clients. However, its interface is highly coupled, which constrains its extensibility and flexibility.
PySyft\cite{ryffel2018syft} and Flower\cite{beutel2020flower} are two other FL frameworks that provide a distributed computing environment for FL. PySyft is a research platform primarily designed for data science applications based on differential privacy. Due to its rapid development cycles, certain versions lack support for FL, not to mention more complex variants of FL paradigms. Flower is a specialized experimental platform for FL that primarily offers users the capability to conduct large-scale FL experiments and explore diverse scenarios involving heterogeneous devices. However, it should be noted that Flower only supports synchronous FL and does not extend to variant paradigms such as asynchronous FL and personalized FL. In addition, these frameworks two lack benchmarks for evaluating the performance of FL algorithms. Additionally, the lightweight framework FLGO \cite{wang2023flgo} has recently garnered considerable attention due to its extensive baseline and benchmark support. However, its reliance on a linear execution method to simulate the FL training process limits its applicability, particularly in time-sensitive scenarios where FLGO fails to offer adequate support.

Although these frameworks have contributed significantly to the field of FL experimentation, they each suffer from certain limitations. Specifically, their lack of scalability and flexibility highlights the urgent need for a more advanced framework. Thus, we propose FedModule, a framework designed with a modular structure and adhering to the "one code, all scenarios" principle.

\section{Framework Design}

\subsection{System Overview}

FedModule consists of two kernels: the Framework Core and the Module Repository. The Framework Core is responsible for creating essential components, such as server, FL benchmark, and message queue, and managing the whole running process. The Module Repository contains various modules, such as updater, scheduler, and mode. Each module can be loaded dynamically to support different federated learning paradigms.

As shown in \cref{fig:system_overview}, the workflow of FedModule is as follows. First, the Framework Core generates the essential components, including server, FL benchmark, message queue, client manager. Then, each component can be assembled and extended according to the user's configuration through the Module Repository to meet different experimental scenarios. After the assembly, the client manager organizes clients to participate in training according to the configuration. It is important to note that, to improve the adaptability of the framework, a variety of methods are available to organize clients, which will be discussed in more detail in \labelcref{section:Custiomize Execution Mode}. During the FL process, the server and the clients communicate through the message queue. This approach decouples the server and the clients, facilitating better extensibility of the framework. In the end, the Framework Core collects the results of the training process and gives the final results to the user.

\subsection{Framework Core \& Module Repository}

In our framework, the Framework Core and Module Repository are essential elements that significantly enhance its flexibility and extensibility. The Framework Core deconstructs the entire federated learning (FL) process into discrete modules, allowing them to be assembled into various FL paradigms much like assembling Lego bricks. Specifically, as shown in Fig. \labelcref{fig:system_overview}, the Framework Core segments the FL process into four primary components: the FL benchmark, message queue, server, and client manager, along with other functional modules stored in the Module Repository. The FL benchmark is responsible for setting up experimental scenarios, including training models, configuring data distribution, and managing client heterogeneity, among other settings. The server and clients correspond to the server and client entities in FL. The Client Manager is a management class designed to control clients, allowing users to set the client’s execution mode and providing interfaces to start and stop clients. This functionality enables the simulation of real-world scenarios where clients may join or leave the network at any time. The message queue acts as middleware facilitating communication between clients and the server. Using a message queue instead of direct point-to-point connections offers two main advantages: it decouples the communication module from the server, simplifying extensions to the communication module; and it abstracts the underlying communication details, allowing users to transmit data through the provided interfaces without dealing with low-level complexities.

When the framework core creates these four components, it sends the user-specified modules to the Module Repository. The Module Repository employs a tool function we developed, called the Module Locator, to locate the required modules. Upon receiving the module paths specified by the user, the Module Locator sequentially loads the relevant packages step-by-step and returns the final required module.

\begin{figure*}[htbp]
    \centering
    \subfigure[The transformation of parallel clients into sequential execution.]{\includegraphics[width=.46\textwidth]{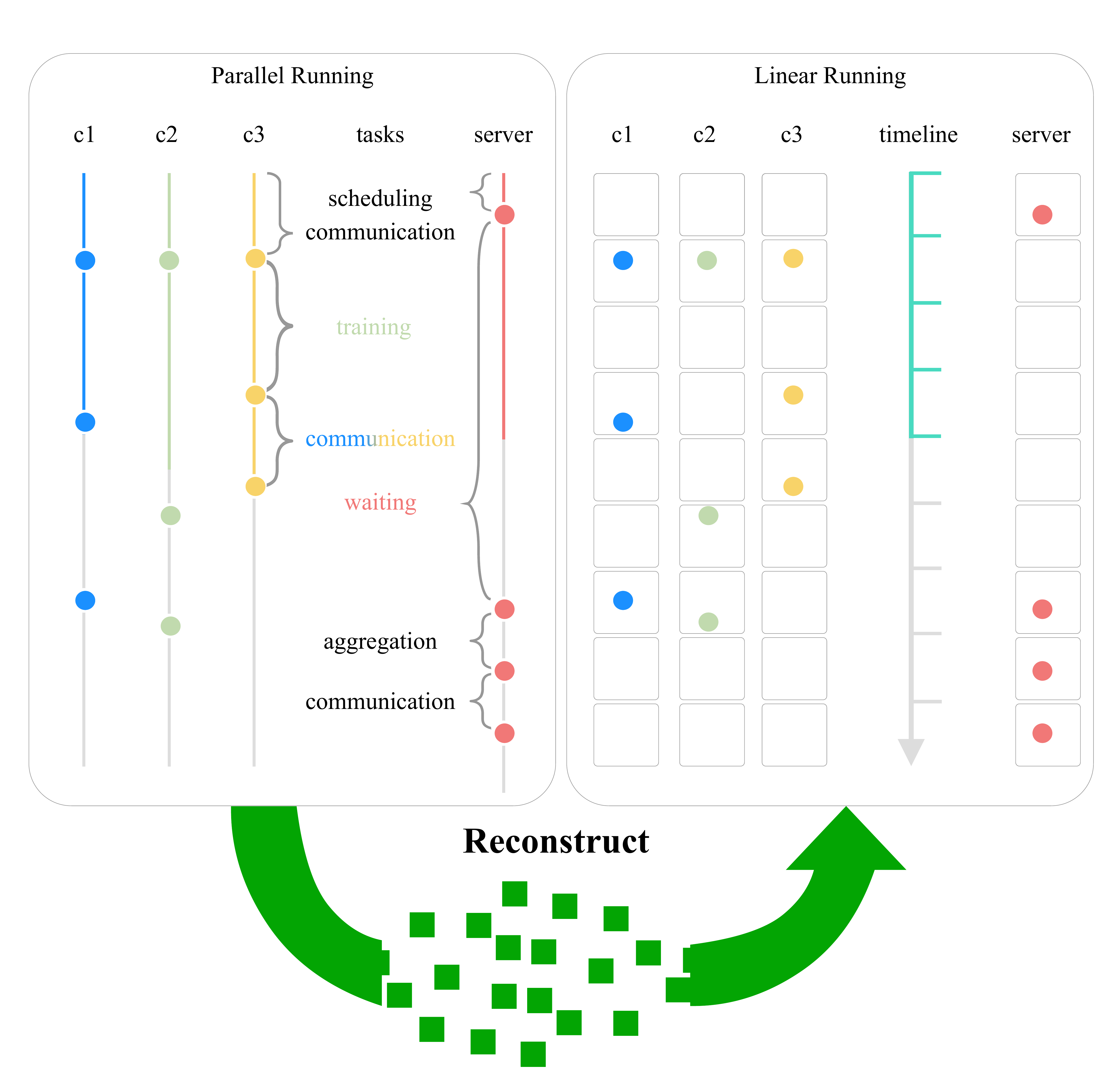}
    \label{fig:timeslice}}
    \subfigure[The Communication framework for distributed mode.]{\includegraphics[width=.42\textwidth]{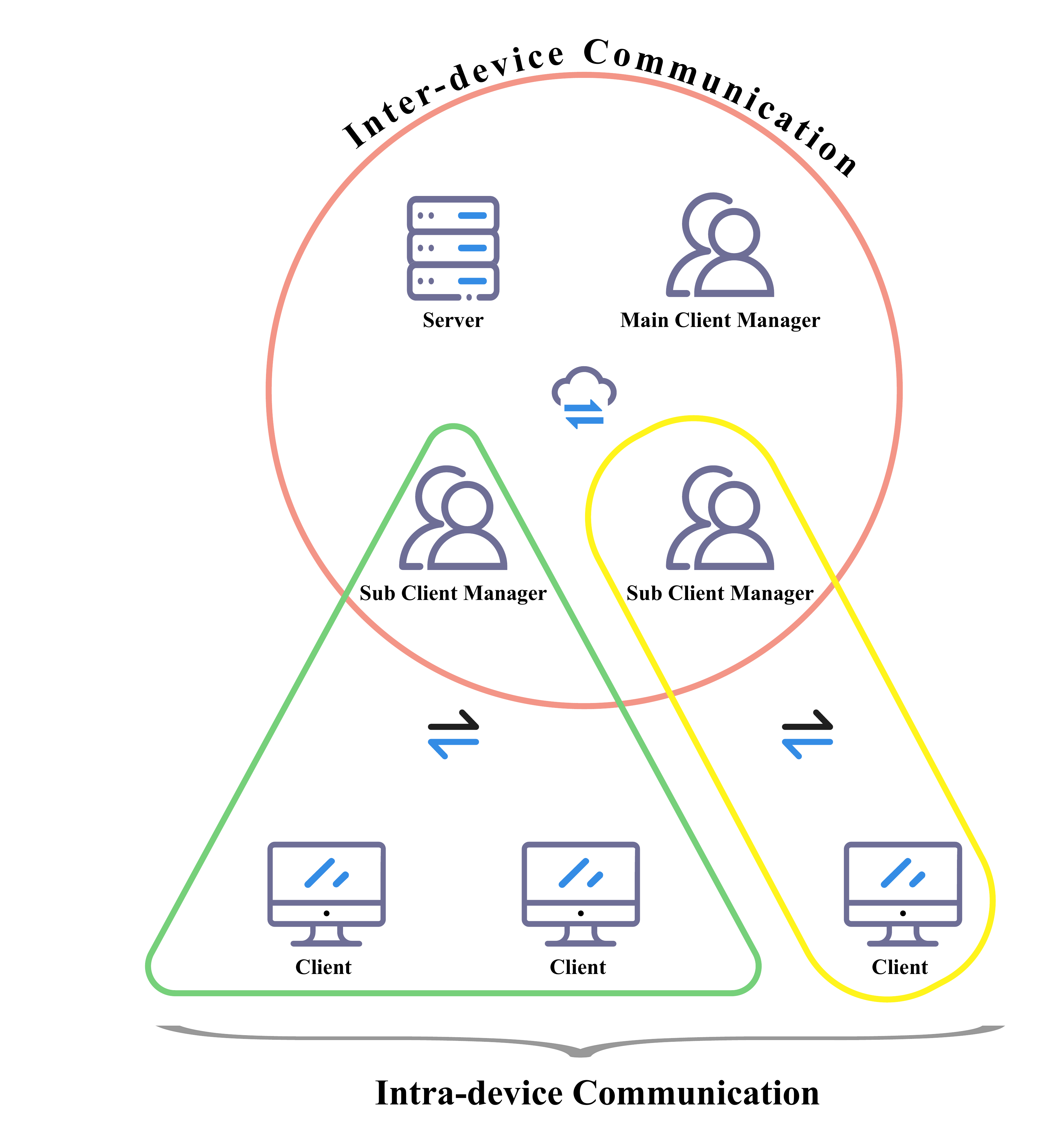}
    \label{fig:distributed}}
    \caption{The illustration of the timeslice mode and distributed mode in FedModule.}
    \label{fig:running_mode}
\end{figure*}

\subsection{Custiomize Execution Mode}\label{section:Custiomize Execution Mode}

To facilitate the slogan of "one code, all scenarios", we make clients to be organized in various ways. Specifically, we utilize the dynamic language feature of Python to implement the execution mode of the clients. Clients can choose their execution mode based on the configuration file. The execution mode can be linear (which means that the clients are running in a linear order, like for loop), thread, process, or even distributed. Despite the variety of execution modes, the client only needs to be implemented once to run in all modes, which is what we advocate as "one code, all scenarios". This design not only simplifies the implementation of the client, but also enhances the flexibility of the framework, making it much easier for researchers to develop.

The implementation of the client follows parallel design principles, inheriting from the thread/process class, with users only needing to implement the run method. Most of the operating modes in our framework are based on processes or threads, which can naturally run such client implementations. However, the linear execution mode cannot be directly supported by the thread/process class, as it requires the clients to run sequentially. Moreover, the distributed mode is more complex than the other modes, as it requires the clients to run on different machines. To address these challenges, we have designed separate solutions specifically for these two modes.

\subsubsection{Timeslice Mode}
Transforming a group of parallel clients into sequential execution is extremely challenging, as it is not possible to inform the clients of each other's execution times in order to enforce a linear order. However, thanks to the dynamic nature of Python, we designed an innovative timeline-based solution, which we called the timeslice mechanism. In order to reconstruct the entire training process to achieve sequential operation, we split the entire FL process into multiple tasks and discretize time into time slices, where a time slice is considered the smallest time unit in the mechanism, and each task corresponds to a different number of time slices. As illustrated in Fig. \labelcref{fig:timeslice}, after reaching a specific time slice, the timeslice mechanism will then proceed to sequentially activate the clients that are required to run at that particular slice. Additionally, it will record the subsequent time slice, which will be utilized to determine the call sequence for each client, based on the feedback provided regarding the running time.

In regard to the conversion of clients running in parallel into discrete tasks that can be invoked by the Timeslice mechanism, we employ the Python's dynamic code modification capabilities and generator functions. In FedModule, the client is composed of numerous task functions, and the tasks themselves are connected by the delay\_simulate function. Therefore, by tracking the usage of the delay\_simulate function within the client, we can split the client into different tasks. However, since tasks utilize shared variables, it is necessary to alter the code's stack space when invoking tasks. This presents a significant challenge, but Python's generator functions offer a solution through the yield keyword, which enables the pausing and resuming of execution in generators. This insight allows us to shift our focus from the initial problem of ``how to split a complete run function into multiple task functions that share a stack but can be invoked independently" to a new problem of ``how to track the delay\_simulate function and convert it and the functions that call it into generator functions." In order to address this issue, it is necessary to perform a static analysis of the syntax tree in order to obtain the function execution chain. Following this, the relevant functions can be converted into generator functions, which will then transform the raw run function into one that meets the requisite linear execution requirements.

\subsubsection{Distributed Mode}

The distributed execution mode permits the execution of users' code on distributed machines, thereby addressing the requirements of large-scale experiments. We run a sub-client manager class on each device, which manages the number of clients running on each device. Furthermore, a main client manager class operates on the server to facilitate the coordination of the sub-client managers and oversee the management of clients on each device. Compared to other execution modes, the communication issues faced by the distributed mode are more complex, so we designed a suitable distributed communication framework for this purpose. As shown in Fig. \labelcref{fig:distributed}, the distributed communication framework is comprised of two main components: the intra-device communication and the inter-device communication. The intra-device communication employs the adapter pattern to wrap the existing communication method, rendering data transmission transparent to the client, whereas the inter-device communication is responsible for the actual communication. The inter-device communication is also modular, with the user able to select the desired communication method, such as socket, MQTT, or HTTP, based on the specific requirements of the experiment.


\subsection{Other Features}
\subsubsection*{Config File}

In contrast to other platforms that employ command-line arguments, FedModule utilizes configuration files for parameter configuration. This approach is advantageous in the context of an evolving FL experimental environment, where the number of required hyperparameters is increasing. Configuration files offer a convenient management and review solution, as well as a more efficient means of reuse and extension. The configuration file is divided into several sections, including the client, server, clientmanager and so on. Each section contains the corresponding hyperparameters, which can be easily modified by the user. The configuration file is loaded by the Framework Core and passed to the corresponding components, which then utilize the hyperparameters to configure the components.

\subsubsection*{DatasetPreLoad Mechanism}

We found that when multiple FL clients are executed in parallel on a single device, the primary limiting factor in client processing speed shifts from computational power to the rate of input/output operations. This is due to the necessity for clients to read data from the disk, which is a time-consuming process that has the potential to significantly impact the overall performance of the FL process. To address this issue, we designed the DatasetPreLoad Mechanism, which is responsible for loading the dataset into memory before clients begin training. The FedModule will create a shared memory space for the dataset which can be accessed by all clients. This approach significantly reduces the time required for the clients to read data from the disk, thereby enhancing the overall performance of the FL process.

\subsubsection*{Abundant Log and Test}

FedModule provides a comprehensive set of diagnostic logs and tests, which facilitates a detailed understanding of the performance of FL algorithms. The logs comprise both online and offline logs, which document the training process and the final results, respectively. The Wandb integration into the framework enables users to visualize the training process online and collect device information.
During the training process, the framework will collect pertinent data, including information regarding data distribution, loss, and accuracy. Furthermore, a comprehensive set of test methods is provided, enabling users to assess the performance of FL algorithms according to a range of metrics. These include detailed accuracy on each class or task, average accuracy of the clients, and other relevant measures.

\begin{table}
    \centering
    \caption{The comparison of different FL frameworks}
    \label{table:FL_Framework_Comparison}
    \begin{tabular}{c|c|c|c|c|c}
        \toprule
        \textbf{Framework} & \textbf{FedModule} & \textbf{TFF} & \textbf{Syft} & \textbf{Flower} & \textbf{FLGo} \\
        \midrule
        \textbf{Scalability} & \ding{51} & \ding{55} & \ding{51} & \ding{51} & \ding{55} \\
        \textbf{Flexibility} & \ding{51} & \ding{55} & \ding{55} & \ding{115} & \ding{115} \\
        \textbf{Benchmark} & \ding{51} & \ding{55} & \ding{51} & \ding{55} & \ding{51} \\
        \textbf{Baselines} & \ding{51} & \ding{55} & \ding{55} & \ding{55} & \ding{51} \\
        \bottomrule
    \end{tabular}
    \\
    \vspace{0.5em}
    \footnotesize{\ding{51} represents that the framework has the corresponding feature, \ding{115} represents that the framework has partial support for the corresponding feature, and \ding{55} represents that the framework does not have the corresponding feature.}
\end{table}

\subsection{FL Framework Comparison}

We compare our framework with other existing FL toolkits, namely TFF, Syft, flower, and FLGo. Table \labelcref{table:FL_Framework_Comparison} gives an overview and a more detailed comparison is provided in the following.

\textbf{Scalability} enables the framework to support large-scale experiments. FedModule offers a diverse set of execution modes to accommodate various experimental requirements and hardware conditions, and supports distributed execution mode, which can run clients on different machines. TFF and FLGo do not support distributed execution mode, which limits their scalability.

\textbf{Flexibility} allows the framework to support various FL paradigms and experimental scenarios. FedModule uses configuration files to define these scenarios, making it easy to customize and adapt paradigms and scenarios as needed. In contrast, TFF, Flower, and Syft support only synchronous federated learning, which limits their flexibility. FLGo runs clients sequentially, which makes it inadequate for time-sensitive experiments.

\textbf{Benchmark} refers to the framework’s ability to provide a comprehensive set of benchmarks for evaluating the performance of FL algorithms. FedModule offers a diverse range of benchmarks, including various datasets and configurable client heterogeneity. In contrast, TFF, Syft, and Flower do not provide benchmarks, which significantly limits their usability. While FLGo does include benchmarks, the variety and number are limited.

\textbf{Baselines} refer to the basic standard FL algorithms that a framework provides. Both FedModule and FLGo offer baselines for comparing the performance of FL algorithms. In contrast, TFF and Syft do not provide baselines, which significantly limits their usability. Flower offers only a limited number of baselines.

\section{Evaluation}

In this section, we conduct experiments to show the ability of FedModule on different federated learning paradigms and benchmarks. We also open-source the config files and code of the experiments to facilitate the reproduction of the results\footnote{code is available at https://github.com/NUAA-SmartSensing/FedModule-Exp}.

\subsection{Experimental Setup}

\subsubsection{Datasets}

In the experiments, we used a total of 4 datasets: CIFAR10 \cite{krizhevsky2009learning}, FashionMNIST \cite{xiao2017fashion}, SVHN \cite{svhn}, and UCIHAR \cite{bulbul2018human}. The CIFAR10 dataset contains 60,000 32x32 color images divided into 10 classes, with 6,000 images per class. It is split into 50,000 training images and 10,000 test images. FashionMNIST includes 60,000 28x28 grayscale images organized into 10 classes, with 6,000 images per class, and is divided into 50,000 training images and 10,000 test images. The SVHN dataset includes 73,257 32x32 color images across 10 classes, with 65,000 images for training and 13,257 for testing. Lastly, the UCIHAR dataset consists of 10,299 instances, split into 7,352 training instances and 2,947 test instances, with each instance having 561 features across 6 classes.

\subsubsection{Training Settings}

Most experiments were conducted on a server equipped with an NVIDIA RTX4090 GPU and an NVIDIA RTX2090 GPU, running Ubuntu 21.04.

Convolutional Neural Networks (CNNs) \cite{cnn} were trained on the FashionMNIST and UCIHAR datasets, while the ResNet-18 architecture \cite{he2016deep} was used for the CIFAR10 and SVHN datasets. Stochastic Gradient Descent (SGD) was employed as the optimizer, with the learning rate set to 0.01. The CNN model utilized in the experiments includes two convolutional layers, two pooling layers, and two fully connected layers. Each selected client undergoes local training for 2 epochs. The batch sizes for the datasets are set to 64 for FashionMNIST, 64 for UCIHAR, and 128 for the remaining dataset. For a comprehensive overview of the hyperparameter settings, please refer to Table \ref{table:Hyper-parameters}. Further details on the hyperparameter configurations for each baseline can be found in our open-source repository.
    
\begin{table*}[htbp]
    \centering
    \caption{Hyper-parameters settings}
    \label{table:Hyper-parameters}
    \subtable[Dataset Related]{
        \begin{tabular}{lcccccc}
            \toprule
            \textbf{Hyperparameter} & \textbf{CIFAR10} & \textbf{EMNIST} & \textbf{FashionMNIST} & \textbf{SVHN} & \textbf{UCIHAR} \\
            \midrule
            Global Epochs & 200 & 200 & 200 & 200 & 150 \\
            Client Numbers & 30 & 30 & 30 & 30 & 30 \\
            Model & ResNet-18 & CNN & CNN & ResNet-18 & CNN \\
            Batch Size & 64 & 64 & 64 & 128 & 128 \\
            Local Epochs & 2 & 2 & 2 & 2 & 2 \\
            Optimizer & SGD & SGD & SGD & SGD & SGD \\
            Learning Rate & 0.01 & 0.01 & 0.01 & 0.01 & 0.01 \\
            \bottomrule
        \end{tabular}
    }
\end{table*}

\subsubsection{Baselines}

We employ the following baseline methods in our experiments: FedAvg \cite{mcmahan2017communication}, FedProx \cite{li2020federated}, FedAdam \cite{fedadam}, FedNova \cite{fednova}, FedAsync \cite{xie2019asynchronous}, TWAFL \cite{chen2019communication}, FedVC \cite{chen2024fedvc}, EAFL \cite{eafl2024}, PFedMe \cite{pfedme}, and FedDL \cite{tu2021feddl}. Specifically, FedProx, FedAdam, and FedNova are utilized in Sections \labelcref{section:Experimental Validation on Diverse Datasets} and \labelcref{section:Abundant Log and Test}; FedAsync, TWAFL, FedVC and EAFL are used in Sections \labelcref{section:Support for Client Heterogeneity} and \labelcref{section:Different FL Paradigms}; and PFedMe, and FedDL are employed in Section \labelcref{section:Different FL Paradigms}.

\subsection{Performance of Different Execution Modes}\label{section:Performance of Different Execution Modes}
\begin{figure}[tbp]
    \centering
    \subfigure[Time]{\includegraphics[width=.48\linewidth]{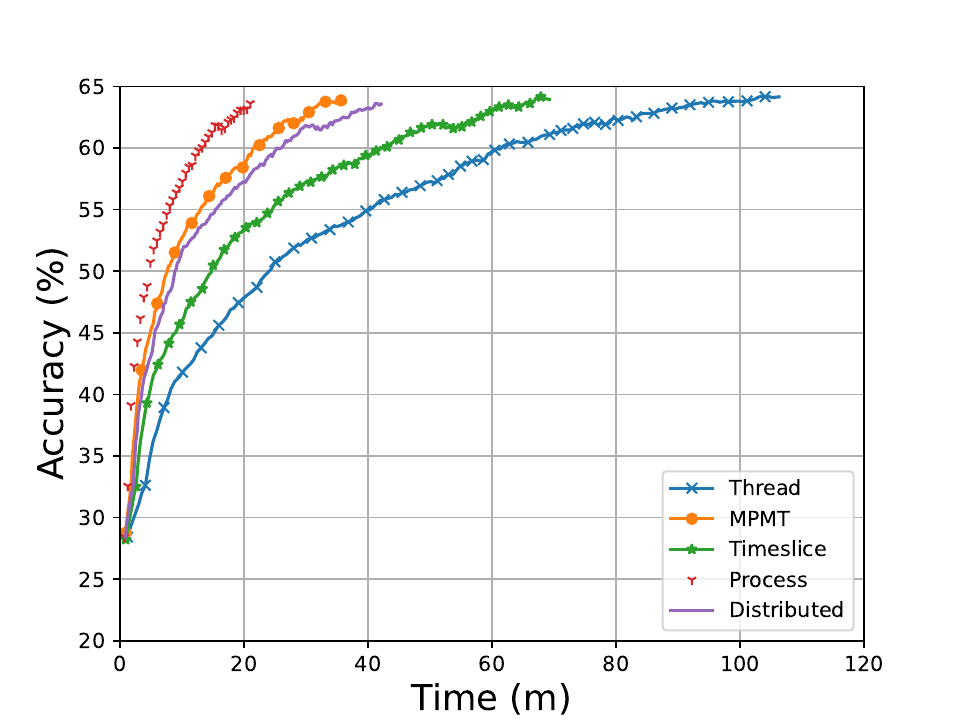}\label{fig:running_mode_performance}}
    \subfigure[Memory]{\includegraphics[width=.48\linewidth]{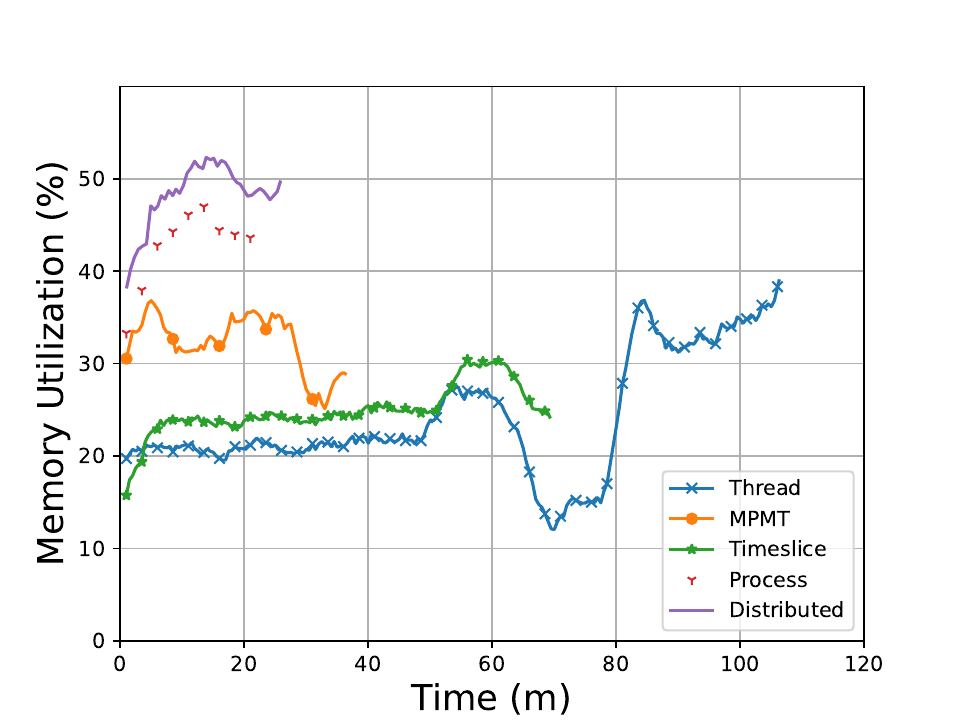}\label{fig:running_mode_mem}}
    \caption{The experiments of execution modes on FashionMNIST in FedModule.}
    \label{fig:running_mode}
\end{figure}

FedModule provides users with the option of selecting different runtime modes, which are designed to accommodate the specific experimental requirements of the user, taking into account the characteristics of the experimental hardware and the available memory. In this section, we evaluate the performance of different execution modes in FedModule. We present the performance of FedAvg using five different execution modes: linear, thread, process, MQMT, and distributed. The dataset employed is CIFAR10. The results are shown in Fig. \labelcref{fig:running_mode_performance}.

As shown in the Fig\labelcref{fig:running_mode_performance}, we used the same random seed for all experiments, resulting in nearly identical accuracy across all excution modes. The execution times for each mode, in ascending order, are as follows: process, mpmt, distributed, timeslice, and thread. The process mode has the shortest execution time because each process operates independently without communication delays between the client and server. The mpmt (multi-process multi-threading) mode, which employs multiple parallel processes to handle client operations, is faster than distributed, timeslice and thread modes. In the distributed mode, additional communication time between devices is introduced compared to the process mode. The thread mode experiences the longest execution time due to the overhead from frequent thread context switching.

Regarding memory consumption, as shown in Fig. \labelcref{fig:running_mode_mem}, the distributed mode has the highest memory usage, followed by process, mpmt, thread, and timeslice. The process mode consumes less memory than the distributed mode because the latter requires additional space to handle inter-device communication. The timeslice mode, due to its non-native sequential execution, requires extra space compared to the thread mode to store the stack information for each client.

The results indicate that FedModule allows users to choose the most appropriate execution mode based on their specific needs, showcasing its flexibility and extensibility, which make it adaptable to various experimental scenarios.

\subsection{Experimental Validation on Diverse Datasets}\label{section:Experimental Validation on Diverse Datasets}
\begin{figure*}[tbp]
    \centering
    \subfigure[CIFAR10]{\includegraphics[width=.24\textwidth]{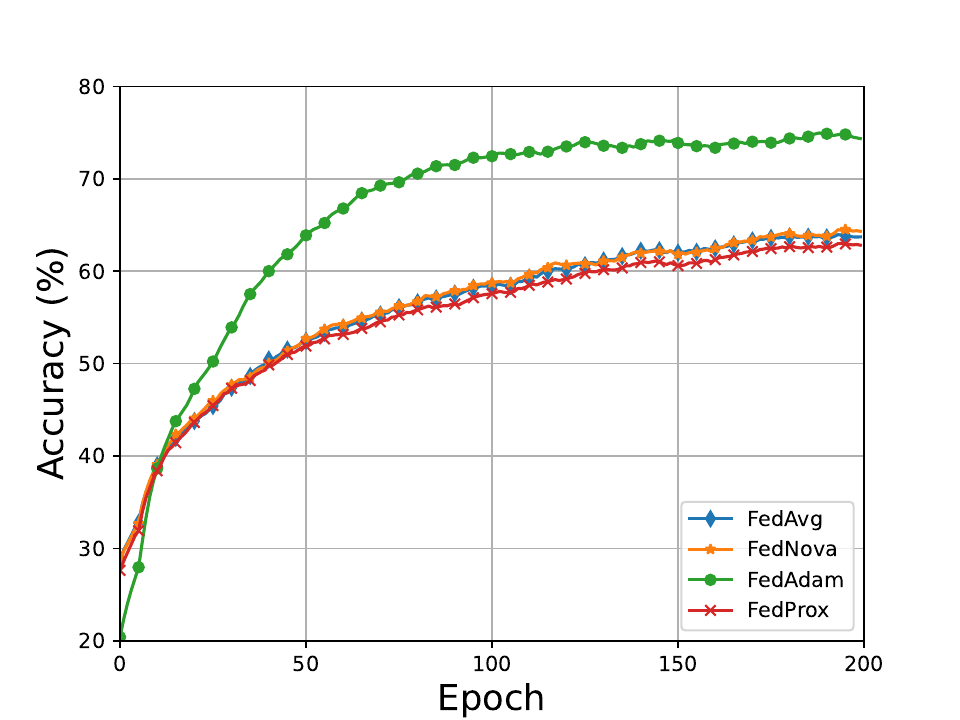}\label{fig:dataset_cifar}}
    \subfigure[FashionMNIST]{\includegraphics[width=.24\textwidth]{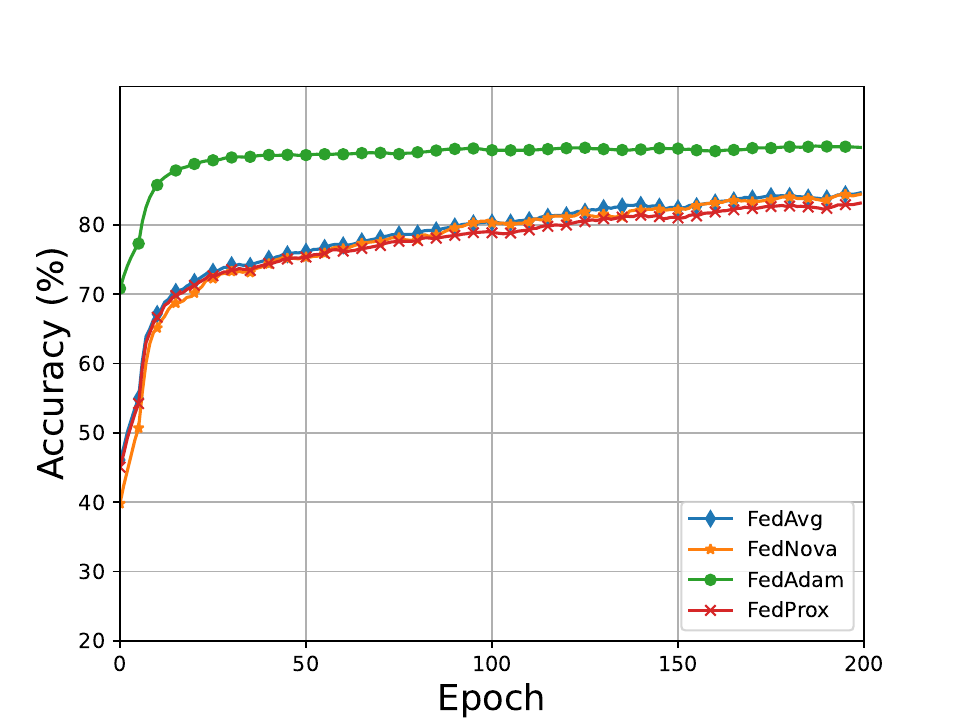}\label{fig:dataset_fs}}
    \subfigure[UCIHAR]{\includegraphics[width=.24\textwidth]{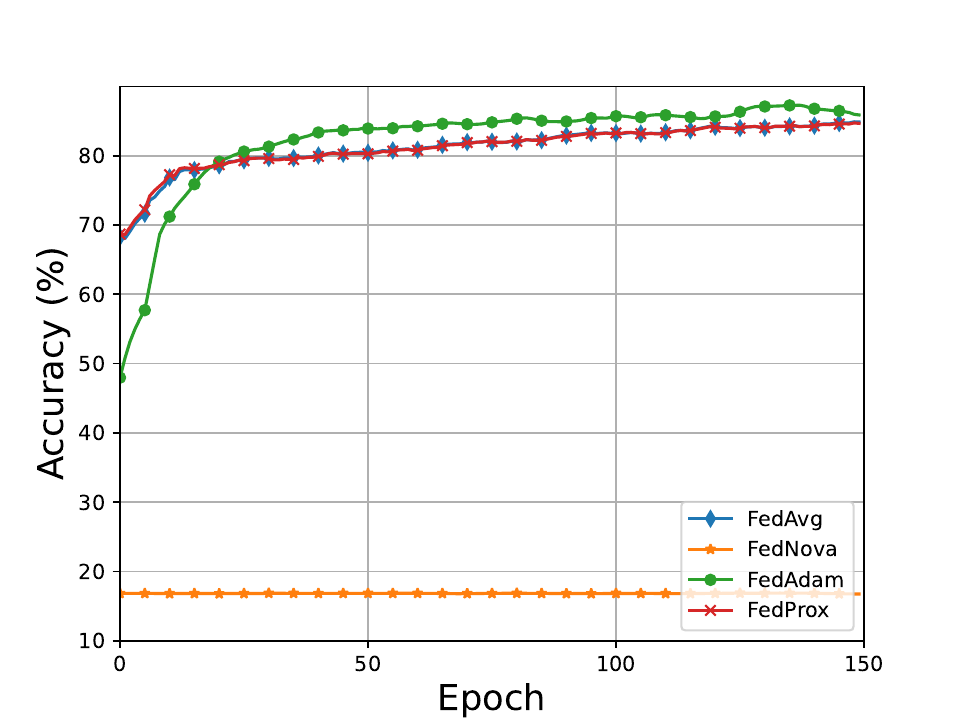}\label{fig:dataset_har}}
    \subfigure[SVHN]{\includegraphics[width=.24\textwidth]{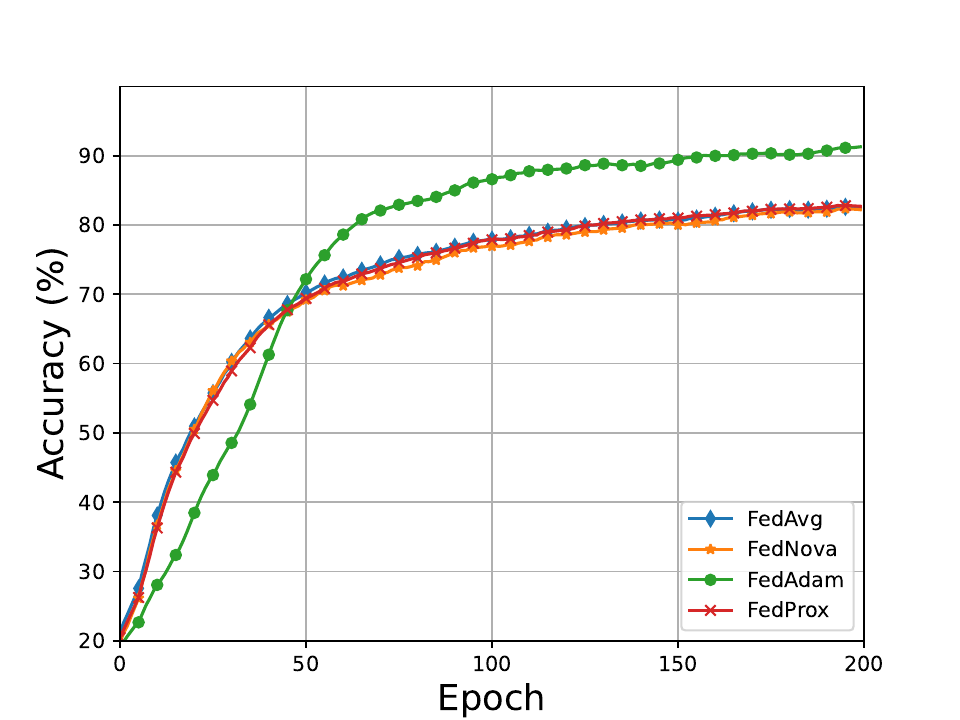}\label{fig:dataset_svhn}}
    \caption{Performance of baselines on different datasets.}
    \label{fig:dataset}
\end{figure*}

Our framework enables seamless switching between different datasets, requiring users to simply configure the appropriate model and parameters without needing to write additional code. We support a wide range of datasets, including not only commonly used datasets but also mixed datasets, streaming datasets, and others. We conducted experiments using several datasets, and the results are presented in Fig. \labelcref{fig:dataset}.

\begin{figure}
    \centering
    \includegraphics[width=0.8\linewidth]{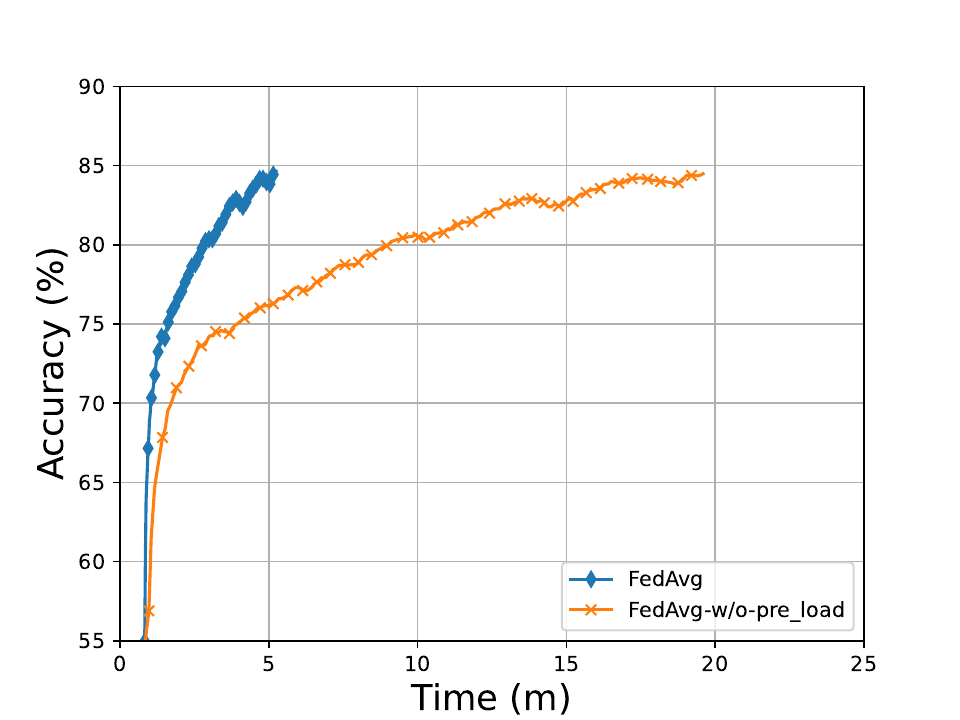}
    \caption{Comparison of time overhead between scenarios with and without dataset preloading.}
    \label{fig:dataset_feature}
\end{figure}
Moreover, FedModule supports loading datasets into memory to enhance experimental speed. We compared the performance with and without dataset preloading, as shown in Fig. \labelcref{fig:dataset_feature}. Experiments on FedAvg in mpmt mode revealed that when clients run concurrently, I/O operations consume a substantial portion of the runtime, thus slowing down the experiments and increasing the divergence between simulation and real-world scenarios. The results show that preloading datasets accelerated the process by 3.6x times compared to loading data during training, validating the effectiveness of this feature and demonstrating its potential to significantly reduce simulation distortions.

\subsection{Support for Client Heterogeneity}\label{section:Support for Client Heterogeneity}
Moreover, FedModule allows the configuration of client heterogeneity to support custom benchmarks. In this section, we will adjust the data heterogeneity and system heterogeneity of the clients to showcase the framework’s capabilities.

\begin{figure}[tbp]
    \centering
    \includegraphics[width=.8\linewidth]{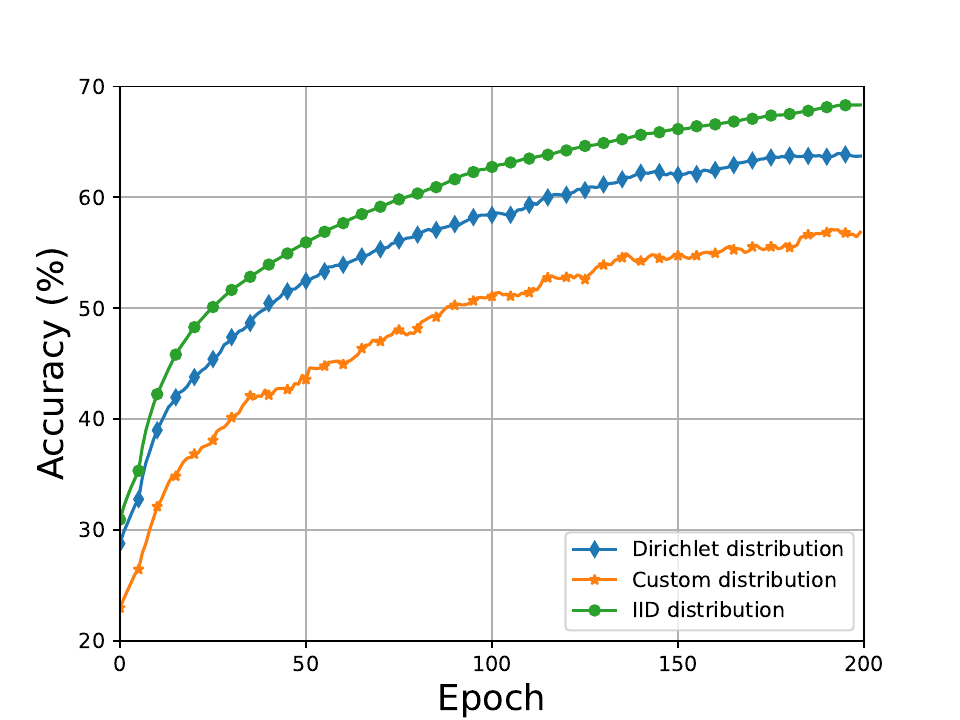}
    \caption{Performance of FedAvg on the FashionMNIST evaluated under three different data distributions.}
    \label{fig:hetero}
\end{figure}

In previous experiments, we used the Dirichlet distribution to configure the data heterogeneity of the clients. In this subsection, we implemented finer-grained configurations by directly specifying the classes and quantities of data each client holds. Of the 30 clients, 10 were assigned data from 3 classes, another 10 were assigned data from 5 classes, and the remaining 10 were assigned data from 7 classes, with all clients maintaining the same total amount of data. Fig. \labelcref{fig:hetero} shows the performance of the FedAvg algorithm across three types of data distributions: independent and identically distributed (i.i.d.), Dirichlet ($\beta=0.5$), and a custom-defined distribution. Fig. \labelcref{fig:log_sample} provides a detailed view of the client-specific data distribution under the Dirichlet distribution ($\beta=0.5$).

\begin{figure}[tbp]
    \centering
    \subfigure[Asynchronous]{\includegraphics[width=.48\linewidth]{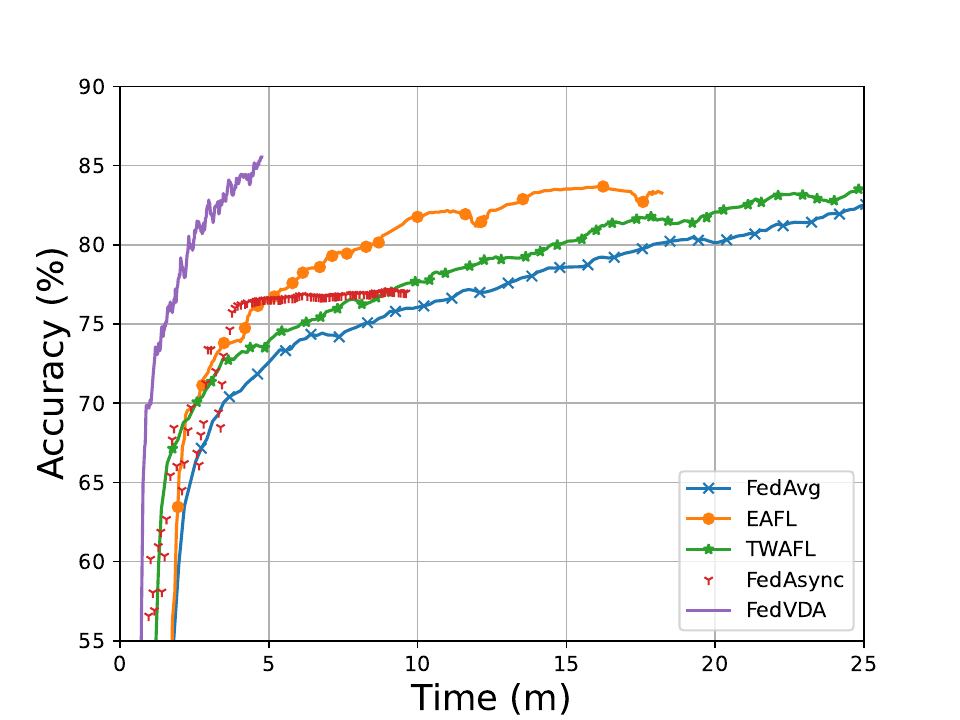}\label{fig:paradigm_async}}
    \subfigure[Personalized]{\includegraphics[width=.48\linewidth]{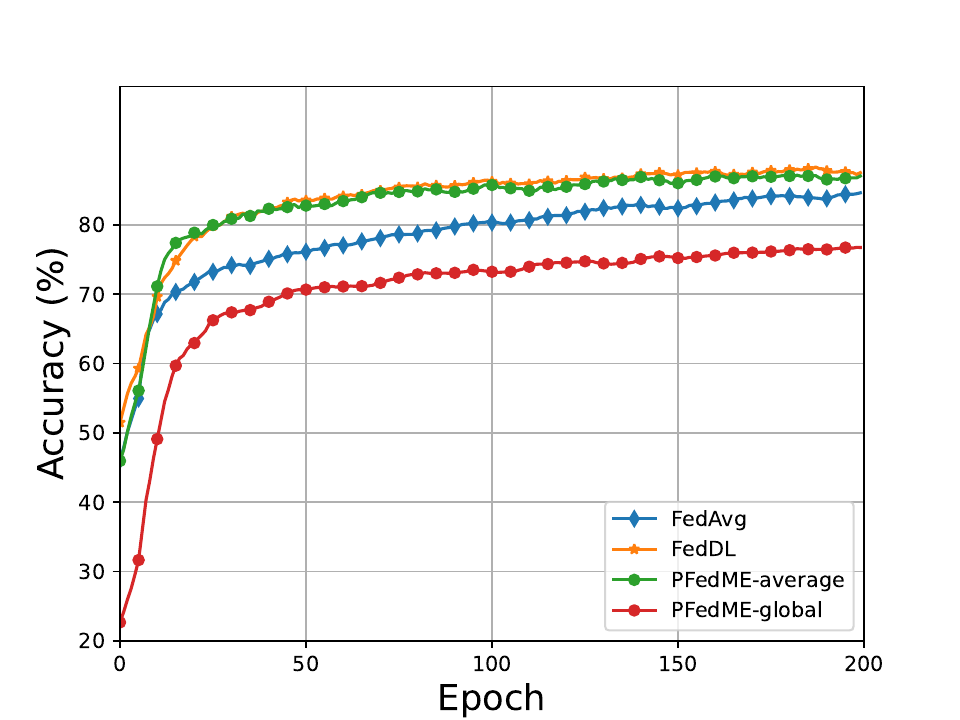}\label{fig:paradigm_personalized}}
    \caption{The illustration of different paradigms on FashionMNIST in FedModule.}
    \label{fig:paradigm}
\end{figure}

In the asynchronous experiments described in Section \labelcref{section:Different FL Paradigms}, we introduced system heterogeneity among the clients. As shown in Fig. \labelcref{fig:paradigm_async}, we configured three types of clients with varying computational speeds: the first type has computational power equivalent to that of a real machine, the second type operates at 20\% of the first type’s computational performance, and the third type operates at 10\% of the first type’s computational performance.
\subsection{Different FL Paradigms}\label{section:Different FL Paradigms}

FedModule supports various FL paradigms, including asynchronous and personalized FL. Our experiments demonstrate the framework's versatility in these paradigms, as shown in Fig. \labelcref{fig:paradigm}. For the asynchronous FL paradigm (Fig. \labelcref{fig:paradigm_async}), we evaluated four different acceleration algorithms: FedAsync, a fully asynchronous algorithm where the server aggregates updates immediately upon receipt; FedVC, a semi-asynchronous algorithm where the server aggregates updates only after collecting a sufficient number; EAFL, another semi-asynchronous algorithm that groups clients and performs asynchronous aggregation within groups and semi-asynchronous aggregation between groups; and TWAFL, a synchronous acceleration algorithm where clients upload only specific model parameters at designated rounds. For the personalized FL paradigm (Fig. \labelcref{fig:paradigm_personalized}), we tested two algorithms: PFedMe, which involves a global model, and FedDL, which aggregates parameters by grouping based on parameter similarity and operates without a global model. These experiments demonstrate the extensive applicability of our framework, accommodating a wide range of federated learning variants. Furthermore, we are actively developing security-related FL paradigms to support experiments focusing on security aspects.

\subsection{Abundant Log and Test}\label{section:Abundant Log and Test}

\begin{figure}[tbp]
    \centering
    \subfigure[Data distribution]{\includegraphics[width=.48\linewidth]{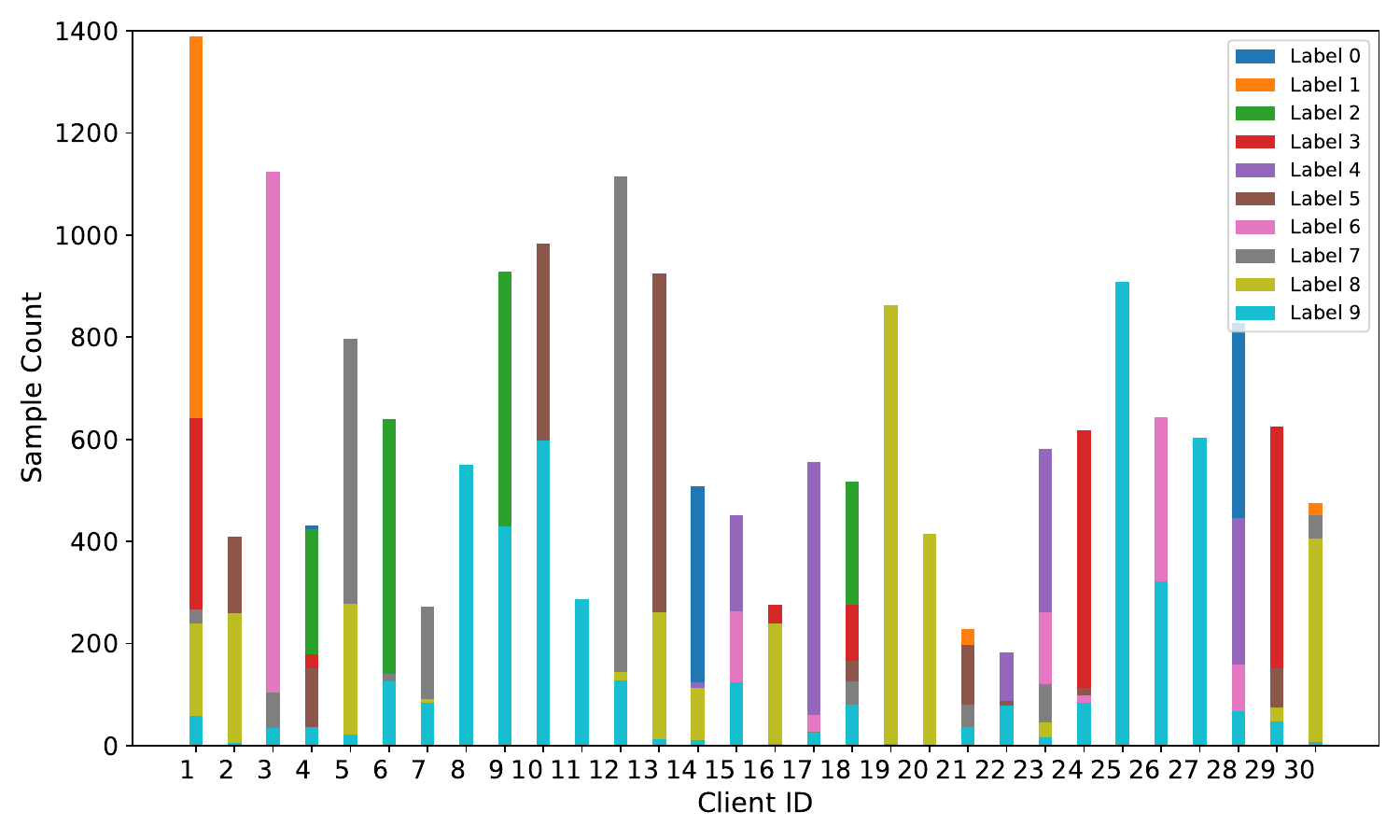}\label{fig:log_sample}}
    \subfigure[Power GPU utilization]{\includegraphics[width=.48\linewidth]{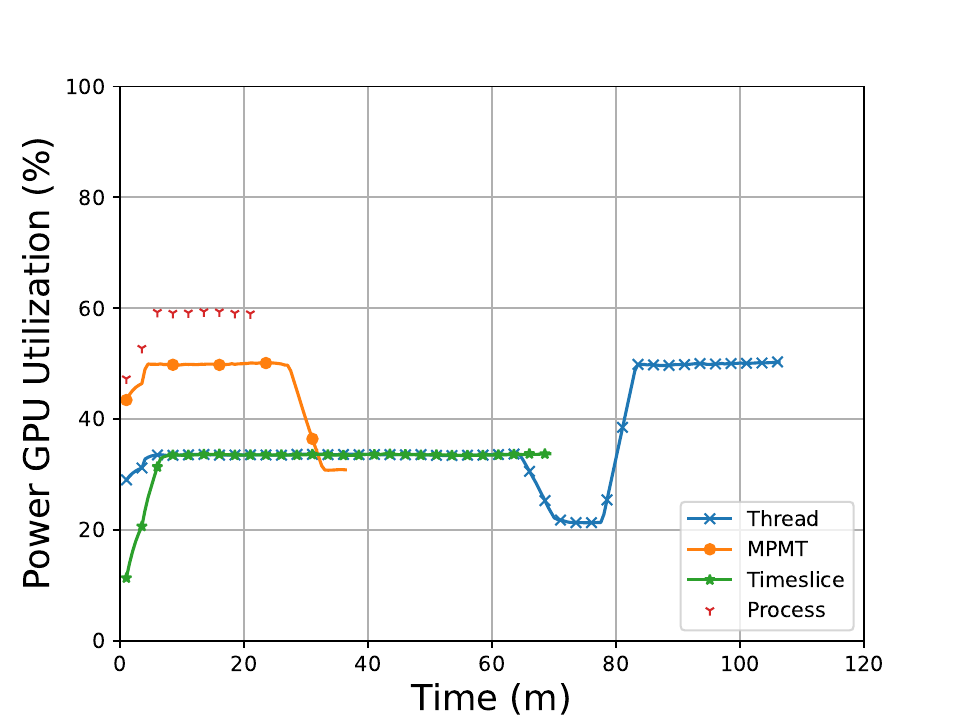}\label{fig:log1}}
    \\
    \subfigure[GPU memory allocated]{\includegraphics[width=.48\linewidth]{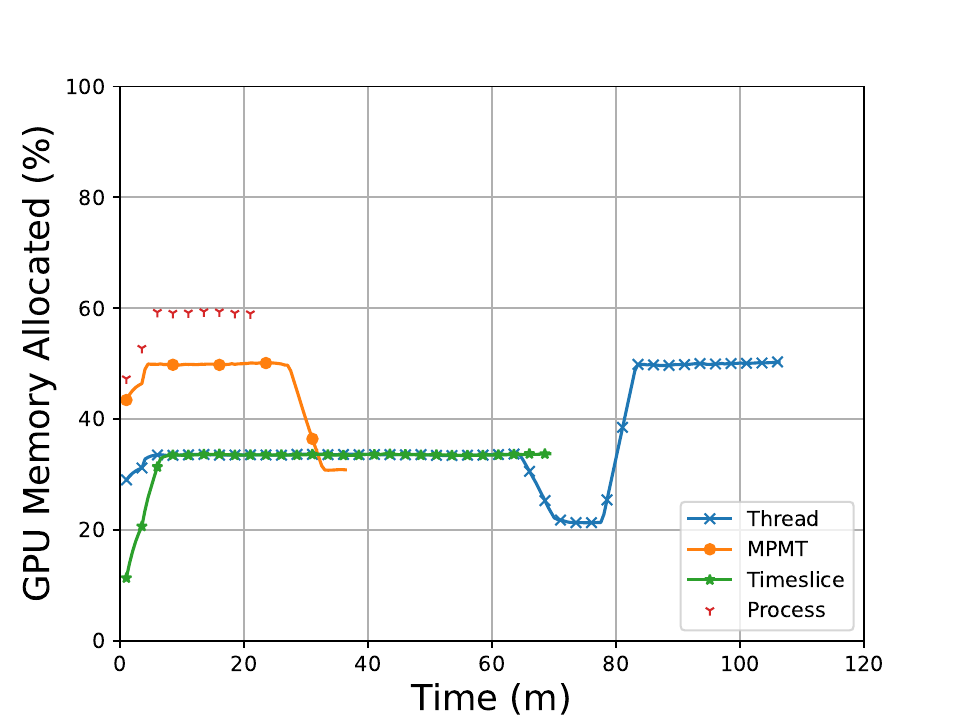}\label{fig:log2}}
    \subfigure[Network traffic]{\includegraphics[width=.48\linewidth]{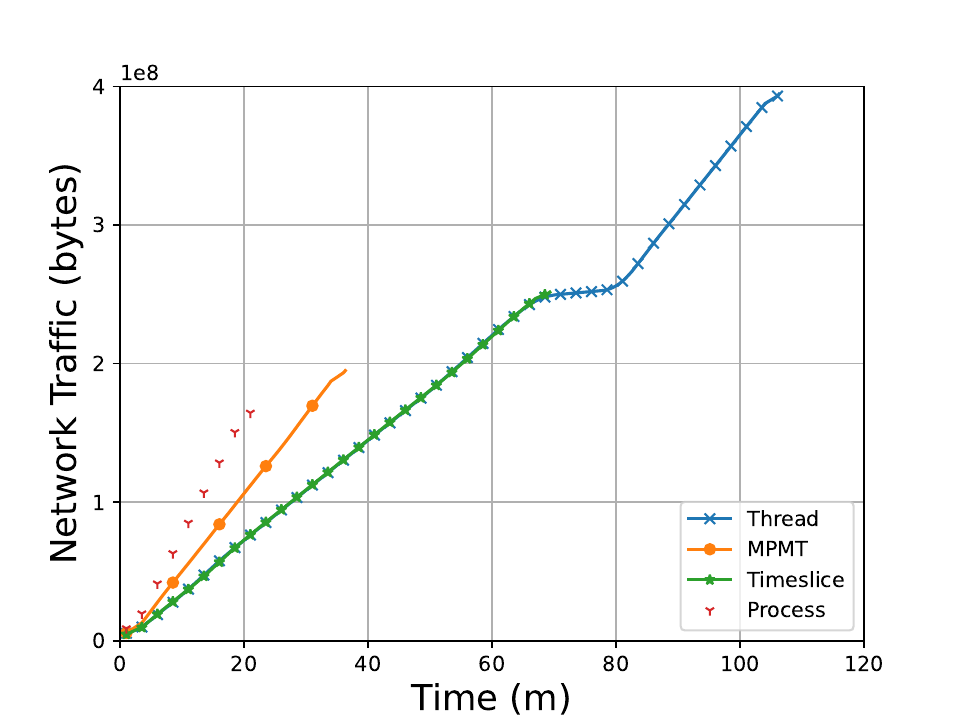}\label{fig:log3}}
    \caption{The illustration of the log and device information in FedModule.}
    \label{fig:log}
\end{figure}

In the previous experimental section, we demonstrated some of the comprehensive data recording capabilities of FedModule, such as tracking test accuracy over time and by logical criteria (Figs. \labelcref{fig:paradigm_async} and \labelcref{fig:dataset_fs}), as well as recording the average accuracy across clients (Fig. \labelcref{fig:paradigm_personalized}). In this section, we present additional experimental records. As shown in the figure, Fig. \labelcref{fig:log_sample} shows the data distribution of the experiment, while Figs \labelcref{fig:log1,fig:log2,fig:log3} depict various device performance metrics during the experiment, such as GPU usage and memory consumption. Additionally, FedModule saves the configuration details of each experiment upon completion and provides an outline summarizing the experiment.

\section{Conclusion}

In this work, we introduce FedModule, a modular FL framework that adheres to the "one code, all scenarios" principle. FedModule decouples the FL training process into multiple independent components, allowing each component to select functional modules based on user requirements to construct specific FL experiments. This modular design enables seamless switching between different FL paradigms and benchmarks. We conducted extensive experiments demonstrating FedModule’s capability to support existing algorithms and validating the effectiveness of its features. In the future, we plan to implement more algorithms within FedModule to further enhance its applicability.

\bibliographystyle{IEEEtran}

\end{document}